\documentclass{bmvc2k}
\usepackage{algorithm}
\usepackage{algorithmic}
\usepackage{amsmath}
\usepackage{caption}

\title{Pseudo-Labeling Curriculum for Unsupervised Domain Adaptation}

\addauthor{Jaehoon Choi}{whdns44@kaist.ac.kr}{1}
\addauthor{Minki Jeong}{rhm033@kaist.ac.kr}{1}
\addauthor{Taekyung Kim}{tkkim93@kaist.ac.kr}{1}
\addauthor{Changick Kim}{changick@kaist.ac.kr}{1}

\addinstitution{
School of Electrical Engineering\\
 Korea Advanced Institute of Science\\
 and Technology (KAIST)\\
 Daejeon, Korea
}

\runninghead{J. Choi, M. Jeong, T. Kim, C. Kim}{Pseudo-Labeling Curriculum}

\def\etal{\emph{et al}\bmvaOneDot}

\begin{document}

\maketitle

\begin{abstract}
To learn target discriminative representations, using pseudo-labels is a simple yet effective approach for unsupervised domain adaptation. However, the existence of false pseudo-labels, which may have a detrimental influence on learning target representations, remains a major challenge. To overcome this issue, we propose a pseudo-labeling curriculum based on a density-based clustering algorithm. Since samples with high density values are more likely to have correct pseudo-labels, we leverage these subsets to train our target network at the early stage, and utilize data subsets with low density values at the later stage. We can progressively improve the capability of our network to generate pseudo-labels, and thus these target samples with pseudo-labels are effective for training our model. Moreover, we present a clustering constraint to enhance the discriminative power of the learned target features. Our approach achieves state-of-the-art performance on three benchmarks: \textit{Office-31}, \textit{imageCLEF-DA}, and \textit{Office-Home}.  
\end{abstract}

\section{Introduction}
\label{Introduction}
In computer vision, large-scale datasets have played an essential role in the success of deep neural networks. However, creating a large amount of labeled data is expensive and time-consuming. To overcome this problem, unsupervised domain adaptation approaches leverage the labeled data in a source domain to train a high-performance model on unlabeled data in a target domain. Such unsupervised domain adaptation methods suffer from the covariate shift between source and target data distributions due to different characteristics of both domains \cite{shimodaira2000}. To tackle the covariate shift problem, many domain adaptation methods aim to jointly learn domain-invariant representations by minimizing domain divergence \cite{ben2010theory}.  

Recently, adversarial domain adaptation methods \cite{Ganin2015, tzeng2017adversarial} have received lots of attention. Similar to generative adversarial networks (GANs) \cite{goodfellow2014generative}, these methods employ two players, namely domain discriminator and feature extractor, to align feature distributions across domains in an adversarial manner. The core module for domain adaptation is domain discriminator, which is trained to discriminate the domain labels of features generated by the feature extractor. Conversely, the feature extractor is trained to provide transferable features that fool the domain discriminator by minimizing the divergence between features generated from the source and the target domain. Although their general efficacy shows impressive results in various tasks like segmentation \cite{hoffman2016fcns,tsai2018learning}, detection \cite{chen2018domain,Kim_2019_CVPR}, and depth estimation \cite{nath2018adadepth}, the adversarial domain adaptation methods suffer from a major problem. Since the domain discriminator only focuses on aligning global domain features, these methods often fail to consider the category information of target samples. Thus, category-level domain alignment enforced by adversarial domain adaptation methods does not guarantee the good target performance due to the lack of categorical target information \cite{Saito_2018_CVPR}. To learn target categorical representation, recent studies \cite{xie2018learning, shu2018a, zhang2018collaborative} utilize pseudo-labels for target samples, thus encouraging a low-density separation between classes \cite{lee2013pseudo}. Although these methods rely on the assumption that the source-trained classifier assigns pseudo-labels to target samples with high confidence, it is difficult to satisfy this assumption especially when domain discrepancy is large. Training the network with false pseudo-labels may degrade the capability of distinguishing discriminant target samples by accumulating classification errors.

\begin{figure}[t]
\begin{center}
\centerline{\includegraphics[height=3.5cm, width=10cm]{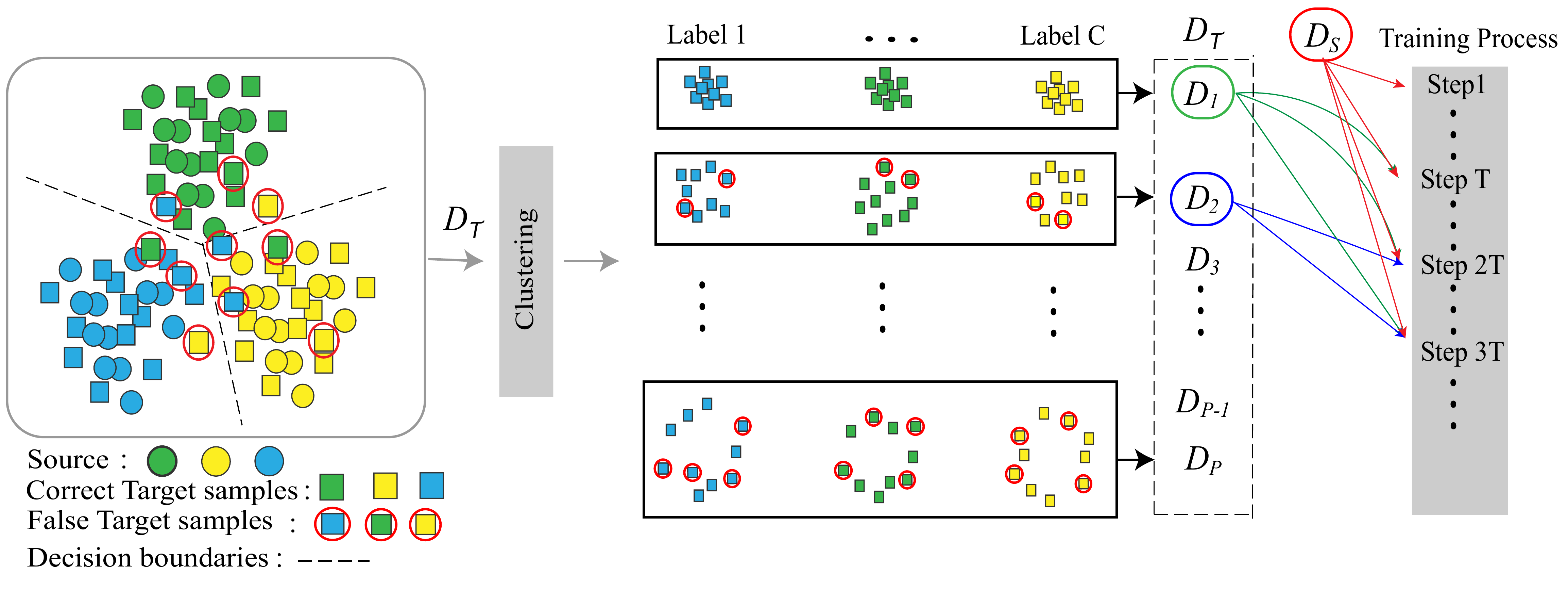}}
\vskip 0.15in
\caption{Learning process of the proposed method. \(D_{\mathcal{S}}\) is a set of all source samples, and \(D_{\mathcal{T}}\) is a set of all target samples. \(D_{1}\), \(D_{2}\), \(\dotsb\), \textrm{and}\: \(D_{P}\) are data subsets obtained from \(D_{\mathcal{T}}\). \(C\) is the number of classes. By using a density-based clustering algorithm, we split our target dataset into a number of data subsets for each category and give these subsets following training process.}
\label{icml-figure1}
\end{center}
\vspace{-6mm}
\end{figure}

To alleviate this issue, we introduce a curriculum learning strategy for unsupervised domain adaptation. The curriculum learning \cite{bengio2009curriculum} is motivated by the cognitive process of humans that gradually learn from easy to complex samples. Inspired by this concept, we propose a novel algorithm called pseudo-labeling curriculum for unsupervised domain adaptation (PCDA). For designing this pseudo-labelling curriculum, we apply a density-based clustering algorithm in \cite{guo2018curriculumnet} to divide target dataset into a number of data subsets relying on both pseudo-labels and features generated by our network. Then, we sequentially train target specific network from target samples whose pseudo-labels have a high chance of being correct to those which are highly likely to be misclassified. Our learning process is demonstrated in Fig. \ref{icml-figure1}. In this process, we only use some target data subsets with correct pseudo-labels at the early training stage. As the training proceeds, our classifier can generate progressively reliable pseudo-labels for other target data subsets which are utilized at the later stage. Training with the pseudo-labeling curriculum boosts the robustness of the training target specific network with pseudo-labels by sequentially providing target data subsets. PCDA is effective to suppress the negative influence of false pseudo-labeled target samples. Furthermore, we suggest the constraints for clustering on the target data to reduce the intra-class variance of target representations and to enlarge the inter-class variance of target representations. This constraint reinforces the assumption of the density-based clustering that target samples with a high density value are more likely to have correct pseudo-labels. Our experimental results suggest that PCDA achieves comparable or even better results compared with the state-of-the-art domain adaptation methods.  


\section{Related Works}
Several types of adversarial learning methods for unsupervised domain adaptation have been shown to match distributions of the features generated from source and target examples (\cite{Ganin2015}, \cite{pmlr-v37-long15}, \cite{bousmalis2016domain}, \cite{pmlr-v70-long17a}, \cite{tzeng2017adversarial}, \cite{Saito_2018_CVPR}). However, they mainly focus on aligning domain-level feature distributions without considering category-level alignment. Xie \etal \cite{xie2018learning} match the labeled source centroid and the pseudo-labeled target centroid to learn semantic representations for target samples. This moving average centroid alignment guides the feature extractor to consider category-level information for target samples. Pei \etal \cite{pei2018multi} exploit multiple domain discriminators to enable fine-grained distribution alignment. 

Some studies \cite{sener2016learning, saito2017asymmetric, shu2018a,french2018selfensembling,zhang2018collaborative} use pseudo-labeled target samples to learn target discriminative representations. Sener \etal \cite{sener2016learning} adopt clustering techniques and pseudo-labels to learn discriminative features. In \cite{shu2018a} and \cite{french2018selfensembling}, they apply a Mean Teacher framework \cite{tarvainen2017mean} to domain adaptation with the consistency regularization. Saito \etal \cite{saito2017asymmetric} employ the asymmetric tri-training (ATT), which leverages target samples labeled by the source-trained classifier to learn target discriminative features. Zhang \etal \cite{zhang2018collaborative} iteratively select pseudo-labeled target samples based on their proposed criterion and retrain the model with a training set including pseudo-labeled samples. However, these methods based on pseudo-labeled target samples have a critical bottleneck that false pseudo-labels can impede learning target discriminative features and further their categorical error is easily accumulated, which hurt the performance. 

Curriculum learning proposed by \cite{bengio2009curriculum} is a learning paradigm based on ranking the training samples from easiest to the most difficult to classify. The curriculum based on ranking is used to guide the order of training procedure. Previous studies (\cite{kumar2010self, lin2018active, jiang2018mentornet, guo2018curriculumnet}) have demonstrated that curriculum learning is useful in handling problems with noisy labels. Guo \etal \cite{guo2018curriculumnet} propose a learning curriculum to handle massive, noisy sets of labels. They measure data complexity using cluster density and split all training samples into a number of subsets ordered by data complexity. By utilizing these subsets, they design a curriculum to train the network. We recast their strategy to design curriculum into our training algorithm.  

\section{Method}
\begin{figure}[t]
\begin{center}
\centerline{\includegraphics[height=5cm, width=10cm]{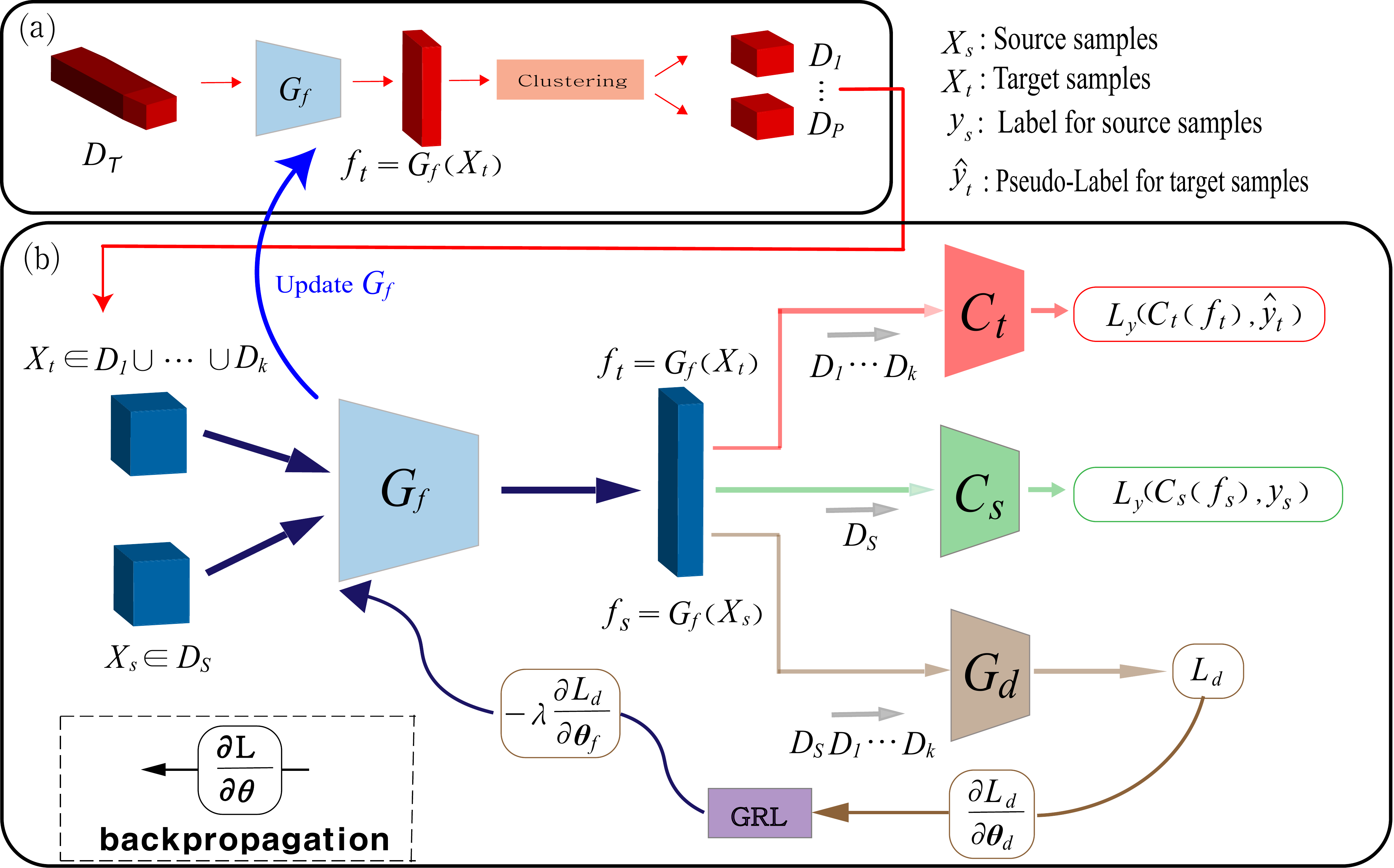}}
\vskip 0.15in
\caption{Pipeline of the proposed method: (a) The curriculum design from density-based clustering, (b) The structure of training our model. In (a),  we use all target samples and pseudo-labels from \(C_{s}\) or \(C_{t}\) to generate data subsets, which form our curriculum for training our model. In (b), we train \(G_{f}\), \(C_{s}\), \(C_{t}\) and \(G_{d}\) with all source samples and selected data subsets of target samples following curriculum.}
\label{icml-figure2}
\end{center}
\vspace{-5mm}
\end{figure}

 \quad In this section, we explain details of the proposed algorithm. Our proposed network structure is demonstrated in Fig. \ref{icml-figure2}. We let \(G_{f}\) denote the shared feature extractor for source and target data. \(C_{s}\) and \(C_{t}\) are classifiers which distinguish features generated from \(G_{f}\). Here, \(\theta_{f}\), \(\theta_{s}\), and \(\theta_{t}\) denote the model parameters of  the respective \(G_{f}\), \(C_{s}\), and \(C_{t}\). \(C_{s}\) learns from the source samples and \(C_{t}\) learns from pseudo-labeled target samples. \(G_{d}\) is the domain discriminator which aims to label all source samples as 1 and all target samples as 0. GRL stands for Gradient Reversal Layer, which inversely back-propagate gradients by changing the sign of the gradient \cite{Ganin2015}. The shared feature extractor \(G_{f}\) is trained from all gradients back-propagated by \(C_{s}\), \(C_{t}\), and \(G_{d}\). Our goal is training \(C_{t}\) to be able to learn discriminative representations for the target domain. We have access to \(N_{s}\) labeled image samples from the source domain \(D_{\mathcal{S}} = \{(x_{i}^s, y_{i}^s)\}_{i=1}^{N_{s}}\) and \(N_{t}\) unlabeled image samples from the target domain \(D_{\mathcal{T}}=\{x_{i}^t\}_{i=1}^{N_{t}}\). \(\hat{y_{i}}^t\) denotes a pseudo-label for the target sample. We assume that \(D_{\mathcal{S}}\) and \(D_{\mathcal{T}}\) completely share the C classes. 

\subsection{Domain Adversarial Learning Scheme}
\label{sec3.1}
\quad Before obtaining pseudo-labeled target samples, we do not train the target specific classifier \(C_{t}\). We only train \(G_{f}\), \(C_{s}\), and \(G_{d}\) by using supervised classification loss and domain adversarial loss following the idea of domain adversarial neural networks (DANN) \cite{Ganin2015}. The overall loss function in this stage is the following:   
\begin{equation} \label{first_stage_loss_function}
\begin{split}
J_{1}(\theta_{f}, \theta_{d}, \theta_{s}) &= \frac{1}{N_{s}} \sum_{i=1}^{N_{s}} L_{y}(C_{s}(f_{i}^s), y_{i}^s)
 - \frac{\lambda}{N_{s}} \sum_{i=1}^{N_{s}} L_{d}(G_{d}(f_{i}^s), d_{i})
 - \frac{\lambda}{N_{t}} \sum_{i=1}^{N_{t}} L_{d}(G_{d}(f_{i}^t), d_{i})\:,
\end{split}
\end{equation}
where \(L_{y}\) is the cross-entropy loss and \(L_{d}\) is the binary cross-entropy loss. \(f_{i}^s\) and \(f_{i}^t\) generated by \(G_{f}\) are the feature representations of source and target samples, \textit{i.e.}, \(f_{i}^s=G_{f}(x_{i}^s)\) and \(f_{i}^t=G_{f}(x_{i}^t)\). \(d_{i} \in \{0, 1\}\) is the domain label and \(\lambda\) is a trade-off parameter. The overall objective is written as follows
\begin{equation} \label{first_stage_optimizer}
\min_{\theta_{f},\,\theta_{s}}\max_{\theta_{d}}\:J_{1}(\theta_{f}, \theta_{d}, \theta_{s}).
\end{equation}
In the early stage of training, since \(C_{s}\) is highly likely to generate false pseudo-labeled target examples, we do not train \(C_{t}\) at all. After a few epochs, we start to train \(C_{t}\) with pseudo-labeled target examples. 

\subsection{Density-Based Clustering for Pseudo-Labeling Curriculum}
\label{sec3.2}
\quad We modify the curriculum learning method from \cite{guo2018curriculumnet} to enable domain adaptation. To design the curriculum, we should address two questions: how to rank the training samples, and how to construct the training schedule. We adopt the ranking method using their clustering density in order to split the target samples. \(C_{s}\) generates pseudo-labels \(\hat{y_{i}}^t\) for target samples and then it separates all features \(f_{i}^t\) into each category. Then, we conduct the following process for each category. We first determine a matrix \(\mathbf{E} \in \rm I\!R^{n \times n}\), where \(n\) is the number of target samples in a certain category. The matrix \(\mathbf{E}\) is composed of elements \(\mathbf{E}_{ij}=\|f_{i}^t-f_{j}^t\|^2\) indicating the Euclidean distance between \(f_{i}^t\) and \(f_{j}^t\). A local density of each sample in a current category is defined as 
\begin{equation} \label{Eq_density_based_clustering}
\rho_{i}=\sum_{j}I(\mathbf{E}_{ij}-e_{c})\:,\enskip
            \:I(x)=\begin{cases}
            1 \quad x < 0\\
            0 \quad \textrm{otherwise}
            \end{cases}.
\end{equation}
We determine \(e_{c}\) by sorting all \(\mathbf{E}_{ij}\) elements in ascending order and select a number ranked at \(k\)\% from the smallest value. We empirically set \(k=40\) in all experiments because the result of clustering is consistent when the value of \(k\) is between 40 and 70. The implication of \(\rho_{i}\) is the number of target samples whose distances are smaller than \(e_{c}\). 

 Target samples with correct pseudo-labels often share similar visual characteristics, and thus features generated from these samples tend to form a cluster for each category. By contrast, target samples with a high chance of being false pseudo-labels often have a visual difference, leading to a sparse distribution which results in a small value of local density. Therefore, we select a sample with the highest local density value as a cluster center for each category. Since target samples close to the cluster center are more likely to have a correct pseudo-label,  we divide target samples into a number of clusters according to their corresponding distances from the cluster center. For each category, we respectively apply the k-means algorithm to the set of distances between the target sample and cluster center for generating \(P\) clusters. Now, we acquire \(P\) clusters in each category, and each cluster can be thought of as a data subset, see Fig. \ref{icml-figure1}. In all our experiments, we empirically set the number of clusters (\(P\)), \textit{i.e.}, \(P\) to 3. Hence, each category contains three subsets, and we combine the overall categories. We can obtain three data subsets \(D_{e}\), \(D_{m}\), and \(D_{h}\), which involve easy, moderate, and hard samples, respectively. The easy samples consist of target samples with a high density value, while the hard samples comprise target samples with a low density value.


In other words, it is challenging for the classifier \(C_{s}\) learned from the source samples to assign correct pseudo-labels for target samples due to the domain gap. In particular, the target samples far from corresponding cluster centers, or distributed near the boundary of the clusters are more likely to have false pseudo-labels. To minimize such an adverse effect, we firstly opt to use the data subset with low intra-class variance and high inter-class variance, such as the easy samples, at the early stage of training. By employing a subset with a high density value, the classifier \(C_{t}\) can learn target discriminative representations from correct pseudo-labeled target samples. 


\subsection{Training with Pseudo-Labeling Curriculum}
\label{sec3.3}
 Our training procedure is proceeded with four stages in conjunction with the ordered sequence of training subsets from easy to hard samples. At the first stage, we train \(G_{f}\), \(C_{s}\), and \(G_{d}\) with \(D_{\mathcal{S}}\) and \(D_{\mathcal{T}}\), while \(C_{t}\) is not trained at all. It is described specifically in Section \ref{sec3.1}. Here \(G_{f}\), \(C_{s}\) and \(G_{d}\) are optimized by Eq. (\ref{first_stage_optimizer}). At the second stage, pseudo-labels for the easy samples \(D_{e}=\{(x_{i}^t, \hat{y_{i}}^t)\}_{i=1}^{N_{e}}\) generated by \(C_{s}\) are highly reliable for training target classifier \(C_{t}\). According to the terminology in Section \ref{sec3.1}, the loss function of the second stage can be written as follows,
\begin{equation} \label{Eq_second_stage_loss_function}
    \begin{split}
     J_{2}(\theta_{f}, \theta_{d}, \theta_{s}, \theta_{t}) &= \frac{1}{N_{s}} \sum_{i=1}^{N_{s}} L_{y}(C_{s}(f_{i}^s), y_{i}^s)
      +\frac{1}{N_{e}} \sum_{i=1}^{N_{e}} L_{y}(C_{t}(f_{i}^t), \hat{y_{i}}^t)\\
      &- \frac{\lambda}{N_{s}} \sum_{i=1}^{N_{s}} L_{d}(G_{d}(f_{i}), d_{i})
      - \frac{\lambda}{N_{e}} \sum_{i=1}^{N_{e}} \beta L_{d}(G_{d}(f_{i}), d_{i})\:,
    \end{split}
\end{equation}
 where \(\beta\) is a hyperparameter that re-weights the contributions of the data subset which is newly added to the training process at the current stage. Compared to the early training samples which are already aligned between the source and target domains, the source and target features generated from newly added samples are poorly aligned. Then, the newly added samples may impede fine-tuning the feature extractor and learning the target classifier. A higher \(\beta\) value encourages the \(G_{f}\) to learn domain invariant features for the newly added samples. Thus, we enable effective domain adaptation based on the curriculum learning by focusing on the contribution of newly added samples. Third, we generate pseudo-labels for the easy samples \(D_{e}\) and the moderate samples \(D_{m}\) by using \(C_{t}\) and also train \(C_{t}\) with \(D_{e}\) and \(D_{m}\). We use the same form of loss function in Eq. (\ref{Eq_second_stage_loss_function}) with adding \(D_{m}\). Lastly, the model is further trained by adding the hard samples \(D_{h}\) according to the learning strategy of the previous stage. Details are included in the supplementary material. We optimize the following minimax objective:
\begin{equation} \label{Eq_second_stage_optimizer}
\min_{\theta_{f},\,\theta_{s},\,\theta_{t}}\max_{\theta_{d}}\:J_{2}(\theta_{f}, \theta_{d}, \theta_{s}, \theta_{t})\:.
\end{equation} 

\subsection{Clustering Constraint}
\label{sec3.4}
A clustering constraint is motivated by the need to boost the effectiveness of the density-based clustering algorithm in order to reduce the number of false pseudo-labeled target samples. For this purpose, the features from target samples with the same category should be as close as possible, while the features from target samples with different categories should be far from each other. We apply the concept of contrastive loss \cite{chopra2005learning}, which takes pairs of examples as input and trains a model to predict whether pairs of inputs are from the same category or not. Unlike \cite{chopra2005learning, schroff2015facenet} which utilize all pairs of images as ground-truth labels, our method is based on the pseudo-labels in order to create pairs of inputs. We propose to formulate the Euclidean-based clustering loss (ECL) as follows, 
\begin{equation} \label{Eq_clustering_constraint}
J_{ECL}(C_{t}) = \sum_{\forall{(i, j)} \in \mathcal{T}} L_{ECL}(h_{i}, h_{j}),\:\:
L_{ECL}(h_{i}, h_{j}) = \begin{cases}
            \|h_{i}^t - h_{j}^t\|^2 & I_{ij} = 1\\
            max(0,\:m - \|h_{i}^t - h_{j}^t\|)^2 & I_{ij} = 0
            \end{cases}\:,\\ 
\end{equation}
 where \(h_{i}^t\) denotes the output of the target classifier \(C_t\) before the softmax layer with respect to the \(i\)-th training sample in the mini-batch. \(\|\cdot\|^2\) is the squared Euclidean distance and \(m > 0\) is a margin. \(I_{ij} = 1\) means that \(h_{i}\) and \(h_{j}\) belong to the same category, while \(I_{ij} = 0\) indicates that \(h_{i}\) and \(h_{j}\) are from the different categories. \(i\) and \(j\) denote the indices of target samples inside the mini-batch. 
 The ECL loss encourages target features from the same category to be closer and enforce the distance between those from the different categories no more than \(m\). This loss can be combined with \(J_{2}\) in Eq. (\ref{Eq_second_stage_loss_function}) to optimize \(G_{f}\) and \(C_{t}\). 
 
\section{Experiments}
\subsection{Setups}
\label{Setups}
\textbf{Office-31} \cite{saenko2010adapting}, the most popular benchmark dataset for visual domain adaptation, contains 4652 images and 31 categories from three distinct domains: \textit{Amazon} (\textbf{A}), \textit{Webcam} (\textbf{W}), and \textit{DSLR} (\textbf{D}). We comprise all domain combinations and build 6 transfer tasks: \textbf{A}\(\rightarrow\)\textbf{W}, \textbf{D}\(\rightarrow\)\textbf{W}, \textbf{W}\(\rightarrow\)\textbf{D}, \textbf{A}\(\rightarrow\)\textbf{D}, \textbf{D}\(\rightarrow\)\textbf{A}, and \textbf{W}\(\rightarrow\)\textbf{A}. 

\noindent\textbf{ImageCLEF-DA}\footnote{https://www.imageclef.org/2014/adaptation}, a benchmark dataset for Image-CLEF 2014 domain adaptation challenge, consists of three domains including \textit{Caltech-256} (\textbf{C}), \textit{ImageNet ILSVRC 2012} (\textbf{I}) and \textit{Pascal VOC 2012} (\textbf{P}). Each domain contains 600 images and 50 images for each category. We evaluate our method on all possible domain adaptation tasks: \textbf{I}\(\rightarrow\)\textbf{P}, \textbf{P}\(\rightarrow\)\textbf{I}, \textbf{I}\(\rightarrow\)\textbf{C}, \textbf{C}\(\rightarrow\)\textbf{I}, \textbf{C}\(\rightarrow\)\textbf{P}, and \textbf{P}\(\rightarrow\)\textbf{C}. Different from Office-31, all images have equal size and the number of images in each category is well-balanced.

\noindent\textbf{Office-Home} \cite{venkateswara2017deep}, a more challenging dataset for domain adaptation evaluation, contains around 15,500 images with 65 categories consisting of four significantly different domains: Artistic images (\textbf{Ar}), Clipart images (\textbf{Cl}), Product images (\textbf{Pr}) and Real-World images (\textbf{Rw}). We can build possible twelve transfer tasks and evaluate our method on all possible tasks. 

\noindent\textbf{Implementation Details.}
The base network for both the baseline and our method is the ResNet-50 \cite{he2016deep}. For domain discriminator, we chose to use the architecture consisting of three fully connected layers with dropout \cite{srivastava2014dropout} and sigmoid layers.  We fine-tuned all feature layers \(G_{f}\) pre-trained on the ImageNet dataset \cite{russakovsky2015imagenet}, following the standard settings for unsupervised domain adaptation \cite{Ganin2015, pmlr-v70-long17a, NIPS2018_7436}. The classifier layers \(C_{s}\), \(C_{t}\) and the domain discriminator layers \(G_{d}\) are trained from the scratch via back-propagation. We set the learning rate of the feature extractor ten times smaller than other layers.  According to the learning rate strategy employed in DANN \cite{Ganin2015}, we utilized mini-batch stochastic gradient descent (SGD) with momentum of 0.9 and the learning rate adjusted by \(\eta = \frac{\eta_{0}}{(1+\alpha p)^\gamma}\), where p is linearly increasing from 0 to 1, \(\alpha=10\), and \(\gamma=0.75\). The total batch size was set to 64, and we selected training samples for each stage in the mini-batch (\(D_{e},\:D_{m},\:D_{h}\)) as follows: (64, 0, 0), (32, 32, 0), and (32, 16, 16) for the stage 1 to 3 in all our experiments. Additionally, we set \(\beta\) to \(2\) or \(3\) in Eq. (\ref{Eq_second_stage_loss_function}) and margin \(m\) to \(2\) in Eq. (\ref{Eq_clustering_constraint}).


\subsection{Evaluation Results}
We complied with standard evaluation protocols for unsupervised domain adaptation \cite{Ganin2015} and used all source samples with labels and all target samples without labels. For all tasks, we performed three experiments and report the averaged results. We compare our methods with the basic ResNet50 \cite{he2016deep}, DAN \cite{pmlr-v37-long15}, RTN \cite{long2016unsupervised}, DANN \cite{Ganin2015}, ADDA \cite{tzeng2017adversarial}, JAN \cite{pmlr-v70-long17a}, GTA \cite{Sankaranarayanan_2018_CVPR}, CAN \cite{zhang2018collaborative}, and CDAN \cite{NIPS2018_7436}. All these methods are based on ResNet50, and results of baselines are directly reported from original papers.  

We report the classification accuracies of the networks in Table \ref{table:office31}, Table \ref{table:imageCLEF}, and Table \ref{table:officehome}. Our proposed PCDA is comparable with the state-of-the-art methods in that it achieves the best average accuracy on the three unsupervised domain adaptation benchmarks. 

\begin{table*}[t]
\begin{minipage}[b]{0.5\textwidth}
\resizebox{\textwidth}{!}{
\centering
\begin{tabular}{l|c|c|c|c|c|c|r}
\hline
\textup{Method} & A\(\rightarrow\)W & D\(\rightarrow\)W & W\(\rightarrow\)D & A\(\rightarrow\)D & D\(\rightarrow\)A & W\(\rightarrow\)A & \textup{Avg} \\ 
\hline\hline
ResNet-50 \cite{he2016deep}    & 68.4$\pm$ 0.2& 96.7$\pm$ 0.1& 99.3$\pm$ 0.1& 68.9$\pm$ 0.2& 62.5$\pm$ 0.3& 60.7$\pm$ 0.3& 76.1\\
DAN \cite{pmlr-v37-long15}    & 80.5$\pm$ 0.4& 97.1$\pm$ 0.2& 99.6$\pm$ 0.1& 78.6$\pm$ 0.2& 63.6$\pm$ 0.3& 62.8$\pm$ 0.2& 80.4\\
RTN \cite{long2016unsupervised}    & 84.5$\pm$ 0.2& 96.8$\pm$ 0.1& 99.4$\pm$ 0.1& 77.5$\pm$ 0.3& 66.2$\pm$ 0.2& 64.8$\pm$ 0.3& 81.6\\
DANN \cite{Ganin2015} & 82.0$\pm$ 0.4& 96.9$\pm$ 0.2& 99.1$\pm$ 0.1& 79.7$\pm$ 0.4& 68.2$\pm$ 0.4& 67.4$\pm$ 0.5& 82.2\\
ADDA \cite{tzeng2017adversarial}    & 86.2$\pm$ 0.5& 96.2$\pm$ 0.3& 98.4$\pm$ 0.3& 77.8$\pm$ 0.3& 69.5$\pm$ 0.4& 68.9$\pm$ 0.5& 82.9\\
JAN \cite{pmlr-v70-long17a} & 85.4$\pm$ 0.3& 97.4$\pm$ 0.2& 99.8$\pm$ 0.2& 84.7$\pm$ 0.3& 68.6$\pm$ 0.3& 70.0$\pm$ 0.4& 84.3\\
GTA \cite{Sankaranarayanan_2018_CVPR} & 89.5$\pm$ 0.5& 97.9$\pm$ 0.3& 99.8$\pm$ 0.4& 87.7$\pm$ 0.5& 72.8$\pm$ 0.3& 71.4$\pm$ 0.4& 86.5\\
CAN \cite{zhang2018collaborative} & 92.5 & 98.8 & 100.0 & 90.1 & 72.1 & 69.9 & 87.2\\
CDAN \cite{NIPS2018_7436} & \textbf{93.1}$\pm$ 0.1& 98.6$\pm$ 0.1& 100.0$\pm$ 0.0& 92.9$\pm$ 0.2& 71.0$\pm$ 0.3& 69.3$\pm$ 0.3& 87.5\\
\hline\hline
\textup{PCDA} (ours, wo ECL) & 90.9$\pm$ 0.4& 98.1$\pm$ 0.1& 100.0$\pm$ 0& 91.0$\pm$ 0.2& 73.3$\pm$ 0.2& 71.5$\pm$ 0.3& 87.5\\ 
\textup{PCDA} (ours) & 92.5$\pm$ 0.3& \textbf{98.7}$\pm$ 0.2& \textbf{100.0}$\pm$ 0& \textbf{93.0}$\pm$ 0.2& \textbf{73.5}$\pm$ 0.3& \textbf{72.5}$\pm$ 0.3& \textbf{88.3}\\
\hline
\end{tabular}
}
\centering
\vskip 0.1in
\caption{Accuracy(\%) on \textit{Office-31}.}\label{table:office31}
\end{minipage}
\hfill
\begin{minipage}[b]{0.5\textwidth}
\centering
\resizebox{\textwidth}{!}{
\begin{tabular}{l|c|c|c|c|c|c|r}
\hline
\textup{Method} & \textup{I}\(\rightarrow\)\textup{P} & \textup{P}\(\rightarrow\)\textup{I} & \textup{I}\(\rightarrow\)\textup{C} & \textup{C}\(\rightarrow\)\textup{I} & \textup{C}\(\rightarrow\)\textup{P} & \textup{P}\(\rightarrow\)\textup{C} & \textup{Avg} \\
\hline\hline
ResNet-50 \cite{he2016deep}    & 74.8$\pm$0.3& 83.9$\pm$0.1& 91.5$\pm$0.3& 78.0$\pm$0.2& 65.5$\pm$0.3& 91.2$\pm$0.3& 80.7\\
DAN \cite{pmlr-v37-long15}    & 74.5$\pm$0.4& 82.2$\pm$0.2& 92.8$\pm$0.2& 86.3$\pm$0.4& 69.2$\pm$0.4& 89.8$\pm$0.4& 82.5\\
DANN \cite{Ganin2015} & 75.0$\pm$0.6& 86.0$\pm$0.3& 96.2$\pm$0.4& 87.0$\pm$0.5& 74.3$\pm$0.5& 91.5$\pm$0.6& 85.0\\
JAN \cite{pmlr-v70-long17a} & 76.8$\pm$0.4& 88.0$\pm$0.2& 94.7$\pm$0.2& 89.5$\pm$0.3& 74.2$\pm$0.3& 91.7$\pm$0.3& 85.8\\
CAN \cite{zhang2018collaborative} & 79.5 & 89.7& 94.7 & 89.9 & \textbf{78.5} & 92.0 & 87.4\\
CDAN \cite{NIPS2018_7436} & 78.3$\pm$0.3& 91.2$\pm$0.2& 96.7$\pm$0.3& \textbf{91.2}$\pm$0.3& 77.2$\pm$0.2& 93.7$\pm$0.3& 88.1\\
\hline\hline
\textup{PCDA} (ours, wo ECL) &78.2$\pm$0.2 & 91.0$\pm$0.3 & 96.5$\pm$0.5 & 90.0$\pm$0.3 & 70.8$\pm$0.3 & 96.0$\pm$0.3 & 87.1\\ 
\textup{PCDA} (ours) &\textbf{80.2}$\pm$0.2 & \textbf{92.5}$\pm$0.3 & \textbf{96.7}$\pm$0.3 & 90.3$\pm$0.1 & 75.2$\pm$0.3 & \textbf{97.0}$\pm$0.1 & \textbf{88.7} \\
\hline
\end{tabular}}
\vskip 0.1in
\caption{Accuracy(\%) on \textit{imageCLEF-DA}.}\label{table:imageCLEF}
\end{minipage}
\end{table*}

\begin{table*}[t]
\resizebox{\textwidth}{!}{\begin{tabular}{l|c|c|c|c|c|c|c|c|c|c|c|c|r}
\hline
\textup{Method} & \textup{Ar}\(\rightarrow\)\textup{Cl} & \textup{Ar}\(\rightarrow\)\textup{Pr} & \textup{Ar}\(\rightarrow\)\textup{Rw} & \textup{Cl}\(\rightarrow\)\textup{Ar} & \textup{Cl}\(\rightarrow\)\textup{Pr} & \textup{Cl}\(\rightarrow\)\textup{Rw} & \textup{Pr}\(\rightarrow\)\textup{Ar} & \textup{Pr}\(\rightarrow\)\textup{Cl} & \textup{Pr}\(\rightarrow\)\textup{Rw} & \textup{Rw}\(\rightarrow\)\textup{Ar} & \textup{Rw}\(\rightarrow\)\textup{Cl} & \textup{Rw}\(\rightarrow\)\textup{Pr} & \textup{Avg} \\
\hline\hline
ResNet-50 \cite{he2016deep} & 34.9& 50.0& 58.0& 37.4& 41.9& 46.2& 38.5& 31.2& 60.4& 53.9& 41.2& 59.9& 46.1\\
DAN \cite{pmlr-v37-long15} & 43.6& 57.0& 67.9& 45.8& 56.5& 60.4& 44.0& 43.6& 67.7& 63.1& 51.5& 74.3& 56.3\\
DANN \cite{Ganin2015} & 45.6& 59.3& 70.1& 47.0& 58.5& 60.9& 46.1& 43.7& 68.5& 63.2& 51.8& 76.8& 57.6\\
JAN \cite{pmlr-v70-long17a} & 45.9& 61.2& 68.9& 50.4& 59.7& 61.0& 45.8& 43.4& 70.3& 63.9& 52.4& 76.8& 58.3\\
CDAN \cite{NIPS2018_7436} & 50.6& 65.9& 73.4& 55.7& 62.7& 64.2& 51.8& 49.1& 74.5& \textbf{68.2}& \textbf{56.9}& 80.7& 62.8\\
\hline\hline
\textup{PCDA} (ours, wo ECL) & 50.7& 72.2& 78.3& 57.2& 68.0& 71.0& \textbf{54.6}& 52.4& 78.3& 67.2& 54.0& \textbf{80.8}& 65.4\\ 
\textup{PCDA} (ours) & \textbf{50.8}& \textbf{73.7}& \textbf{78.5}& \textbf{58.6}& \textbf{70.4}& \textbf{71.2}& 53.9& \textbf{53.5}& \textbf{79.0}& 65.6& 53.6& 80.2& \textbf{65.8}\\
\hline
\end{tabular}}
\vskip 0.1in
\caption{Accuracy(\%) on \textit{Office-Home}.}\label{table:officehome}
\end{table*}

\begin{table}[t]
\begin{minipage}[b]{0.3\linewidth}
\centering
\resizebox{\textwidth}{!}{\begin{tabular}{l|c|c|c|c}
\hline
 & A\(\rightarrow\)W &  A\(\rightarrow\)D &  D\(\rightarrow\)A &  W\(\rightarrow\)A \\
\hline\hline
\(\beta\)=1 & 87.8 & 89.4 & 70.8 & 71.0\\
\(\beta\)=2 & 90.3 & 89.0 & \textbf{73.3} & \textbf{71.5} \\
\(\beta\)=3 & \textbf{90.9} & \textbf{91.0} & 70.9 & 71.1 \\
\hline
\end{tabular}}
\vskip 0.1in
\caption{The effectiveness of \(\beta\).}
\label{table:beta}
\end{minipage}%
\hfill
\begin{minipage}[b]{0.2\linewidth}
\centering
\resizebox{\textwidth}{!}{\begin{tabular}{l|c|c}
\hline
 & D\(\rightarrow\)A &  Ar\(\rightarrow\)Pr \\
\hline\hline
\(P\)=2 & 70.5 & 47.6  \\
\(P\)=3 & \textbf{73.5} & 50.8 \\
\(P\)=4 & 72.8 & \textbf{51.3}\\
\hline
\end{tabular}}
\vskip 0.1in
\caption{The effectiveness of \(P\).}
\label{table:numberofcluster}
\end{minipage}%
\hfill
\begin{minipage}[b]{0.42\linewidth}
\centering
\resizebox{\textwidth}{!}{\begin{tabular}{l|c|c|c|c}
\hline
 & D\(\rightarrow\)A &  W\(\rightarrow\)A &  I\(\rightarrow\)P & Cl\(\rightarrow\)Rw \\
\hline\hline
\textbf{Model-1} & 70.4 & 70.3 & 76.2 & 70.6 \\
\textbf{Model-2} & 71.8 & 71.4 & 77.3 & 71.0 \\
\textbf{Model-3} & \textbf{73.3} & \textbf{72.5} & \textbf{80.2} & \textbf{71.2} \\
\hline
\end{tabular}}
\vskip 0.1in
\caption{Test accuracy of three different models.}
\label{table:CurriculumLearningStrategy}
\end{minipage}
\end{table}


\subsection{Analysis}

\textbf{Ablation Study.}
We perform an ablation study to check the efficacy of each component in our work. First of all, the above Tables \ref{table:office31}, \ref{table:imageCLEF}, and \ref{table:officehome} show the results of PCDA (wo ECL) and PCDA. PCDA (wo ECL) denotes our training method without clustering constraint. The performance gap between PCDA and PCDA (wo ECL) in \textit{Office31} and \textit{imageCLEF-DA} demonstrate the effectiveness of the clustering constraint. However, regarding \textit{Office-Home} experiment, PCDA with clustering constraint fails to enhance the classification performance in some tasks such as Rw\(\rightarrow\)Cl and Rw\(\rightarrow\)Ar. This is presumably because the training method does not have sufficient mini-batch size to guarantee that similar pairs are somewhat sampled.

\noindent\textbf{Hyperparameter Study.}
Our approach sets the hyperparameter \(\beta\) to learn domain-invariant features from newly added target samples. If \(\beta\) is one, our loss function for adversarial adaptation learning becomes equivalent to DANN's loss function in Eq. (\ref{first_stage_loss_function}). We used the \textit{Office-31} dataset to validate the effectiveness of \(\beta\) in Table \ref{table:beta}. When \(\beta\) is one, the performance of our training process degrades. In contrast, when we set \(\beta\) between 2 or 3, our method achieves better performance. This results prove our assumption that the weight \(\beta\) helps the network to extract domain invariant features for the newly added target samples.    

The number of clusters \(P\) is important hyperparameter for our pseudo-labeling curriculum. When \(P\) is set to 4, We observe that most tasks in \textit{Office-Home} shows performance improvement, while some tasks in \textit{Office-31} present the performance degradation shown in Table \ref{table:numberofcluster}. This is because large \(P\) is effective to improve models that tend to generate relatively many incorrect pseudo-labels for target samples. On the contrary, large \(P\) may not be helpful to enhance the capability of models which already show good performance on target data. We set \(P\) to 3 because the pseudo-labeling curriculum becomes simple to implement and achieves the best performance on all benchmarks.   


\begin{figure*}[t]
    \centering
    \subfigure[Easy samples \(D_{e}\)]{\includegraphics[width=0.24\textwidth]{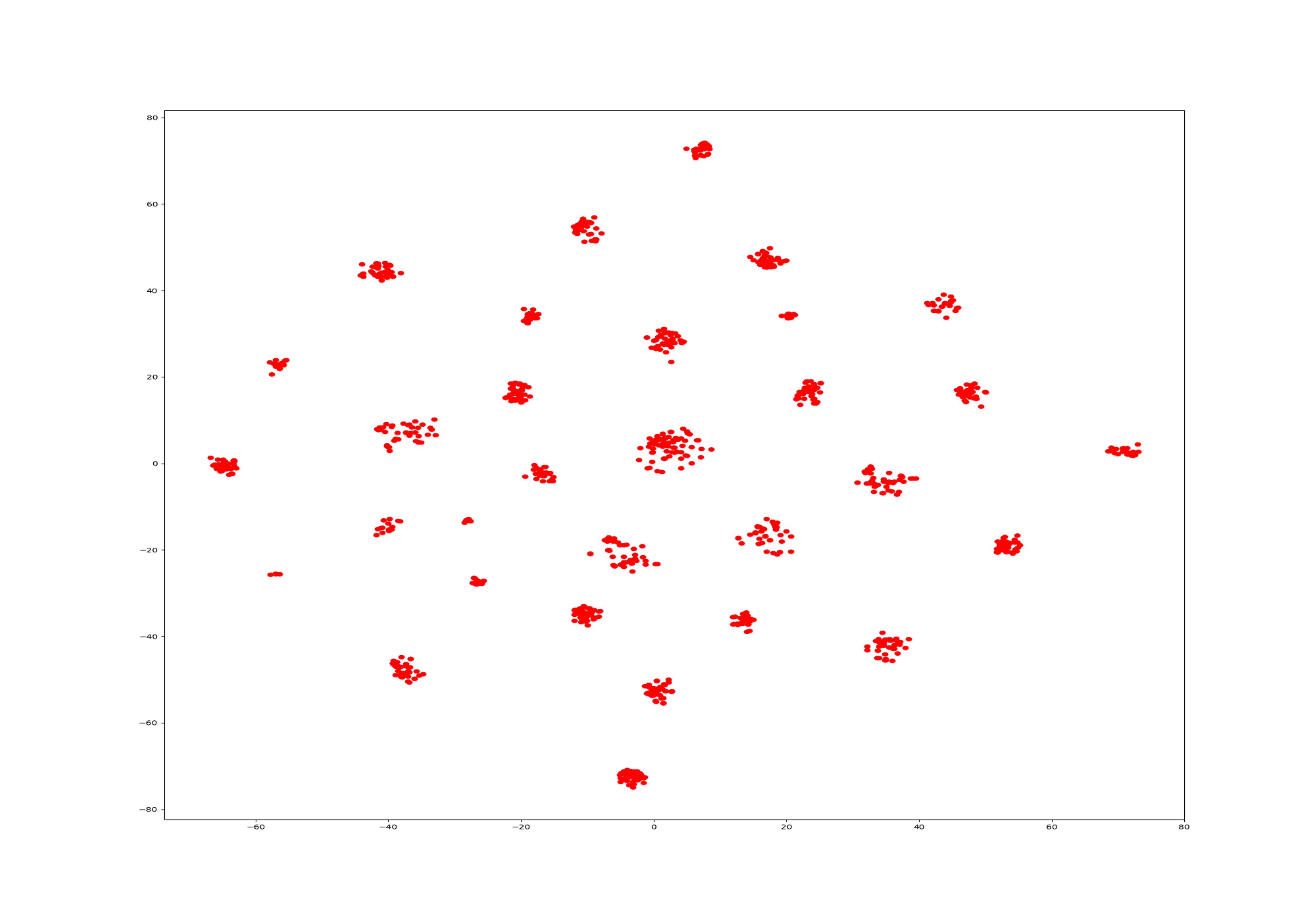}\label{fig:Easy}}
    \subfigure[Moderate samples \(D_{m}\)]{\includegraphics[width=0.24\textwidth]{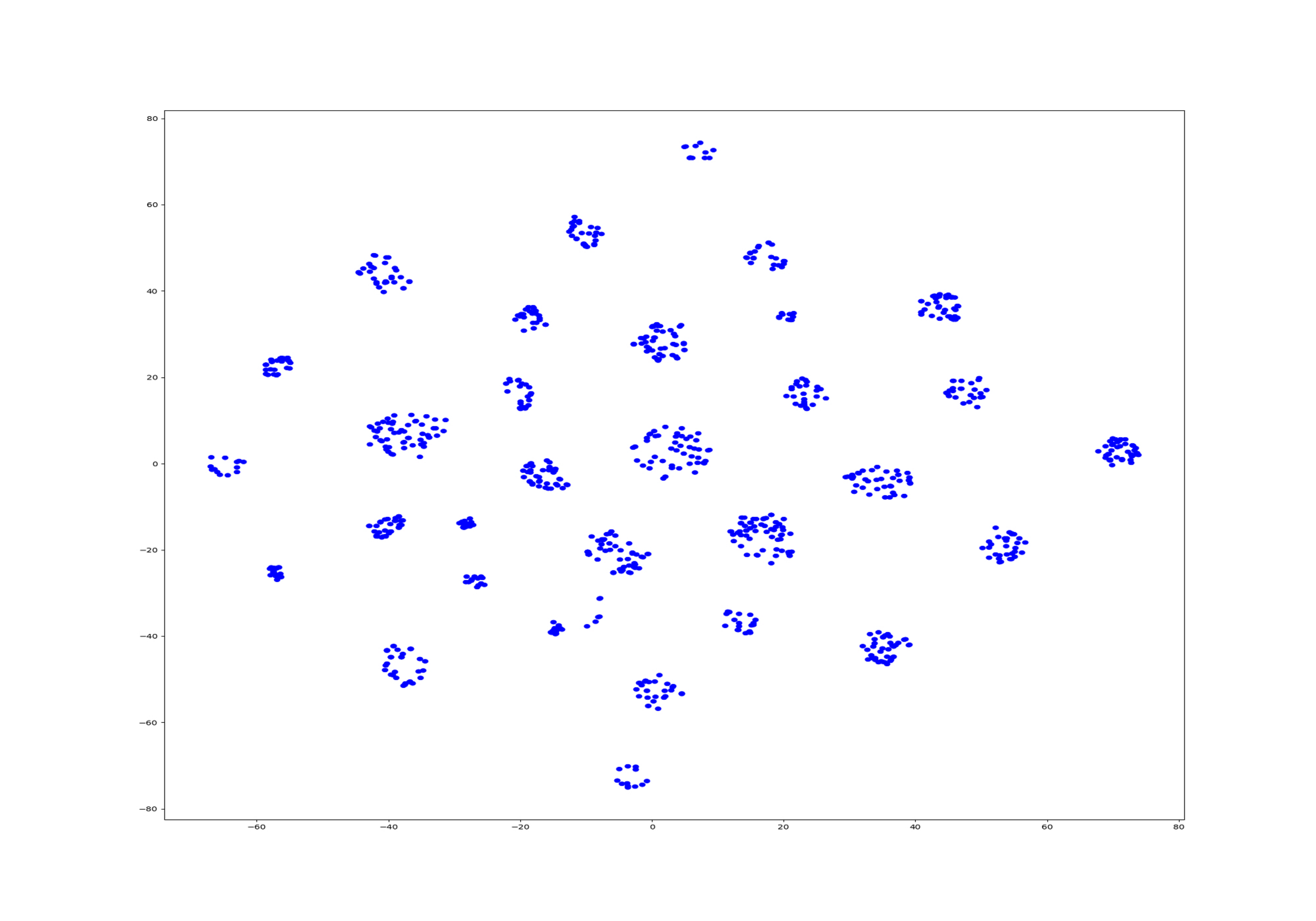}\label{fig:Moderate}}
    \subfigure[Hard samples \(D_{h}\)]{\includegraphics[width=0.24\textwidth]{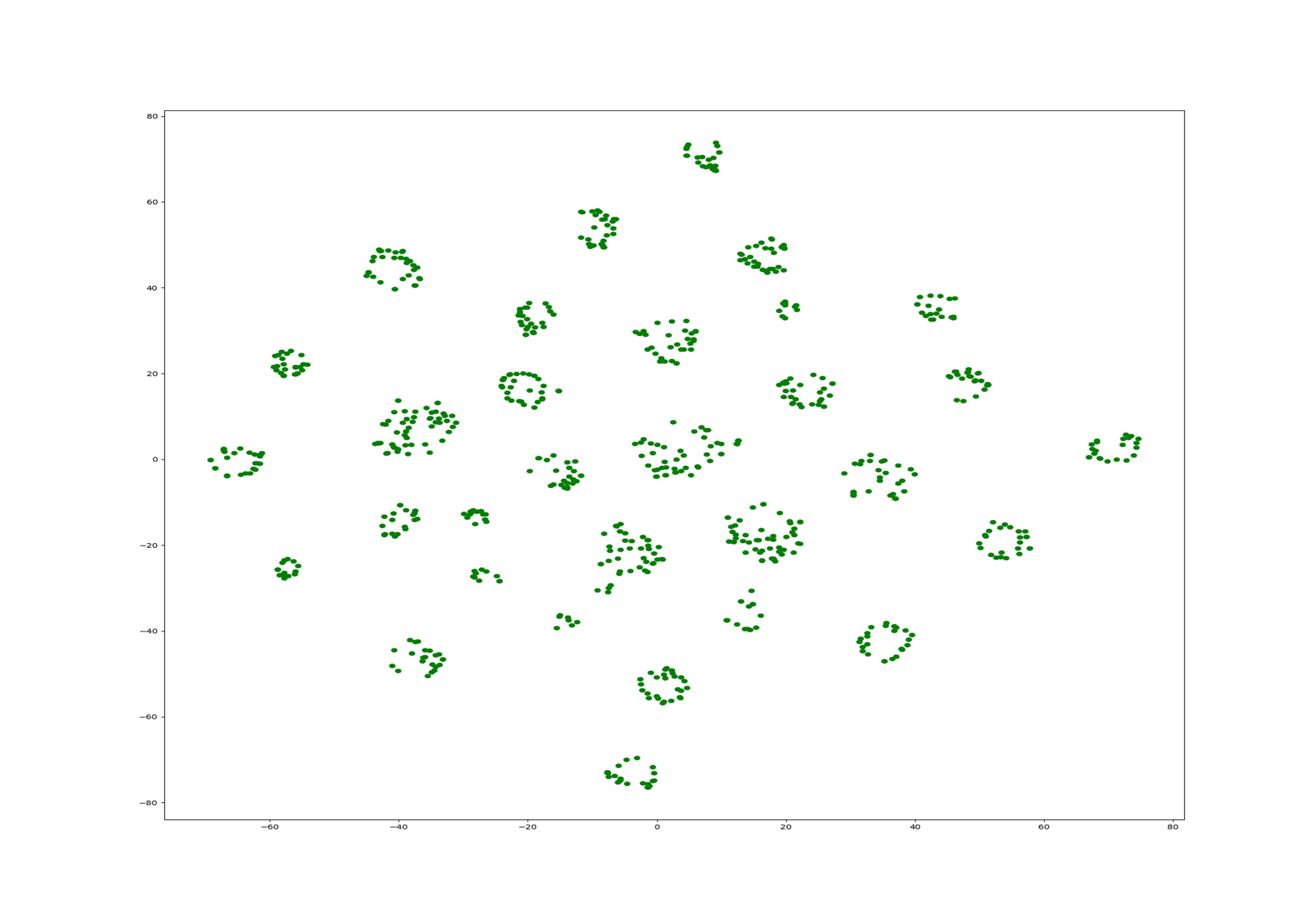}\label{fig:Hard}}
    \subfigure[All target samples \(D_{\mathcal{T}}\)]{\includegraphics[width=0.24\textwidth]{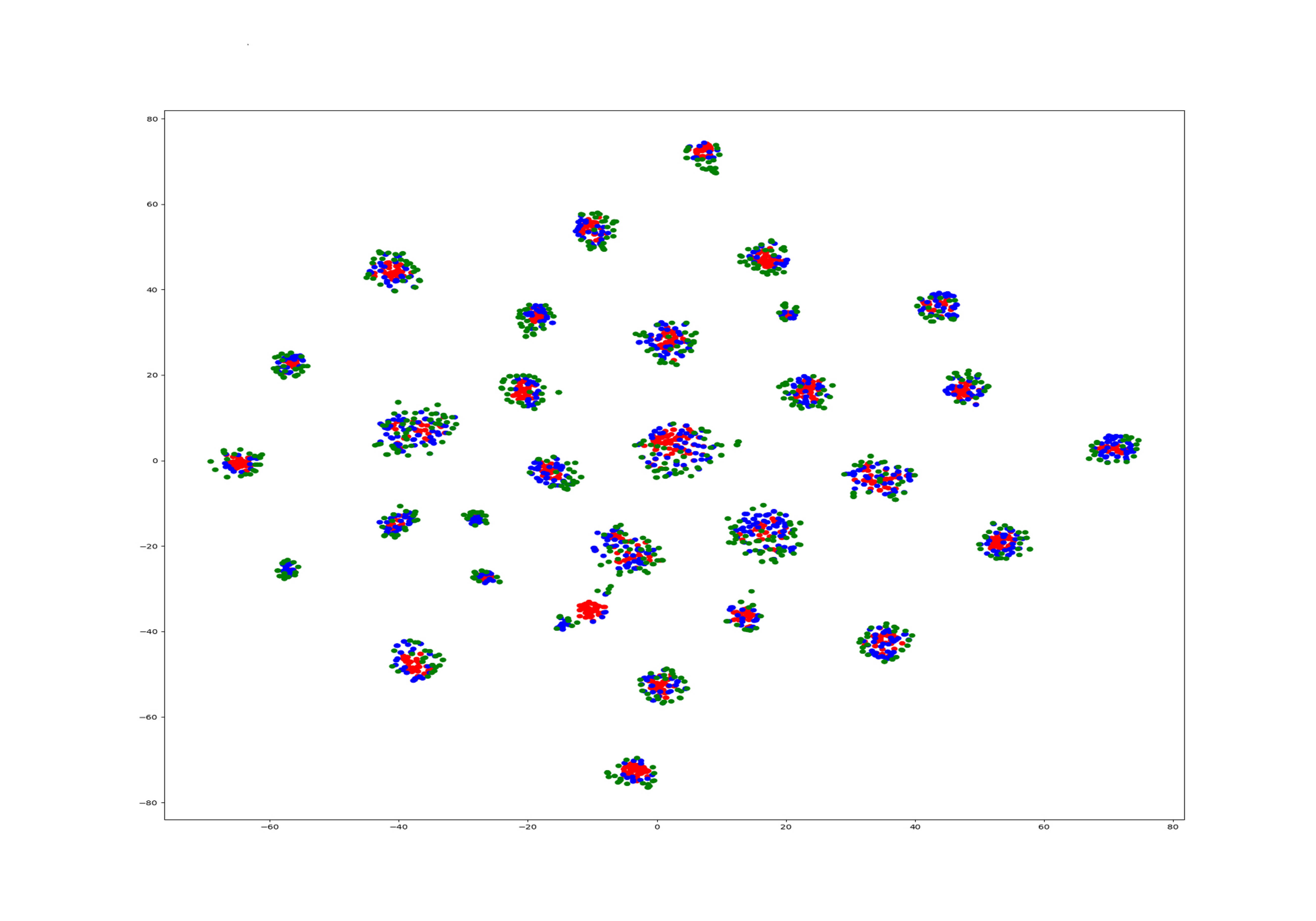}\label{fig:All}}
    \vskip 0.1in
    \caption{T-SNE visualization of features from easy, moderate, and hard samples by using our PCDA method on the D\(\rightarrow\)A task.}
    \centering
    \subfigure[DANN (D\(\rightarrow\)A)]{\includegraphics[width=0.24\textwidth]{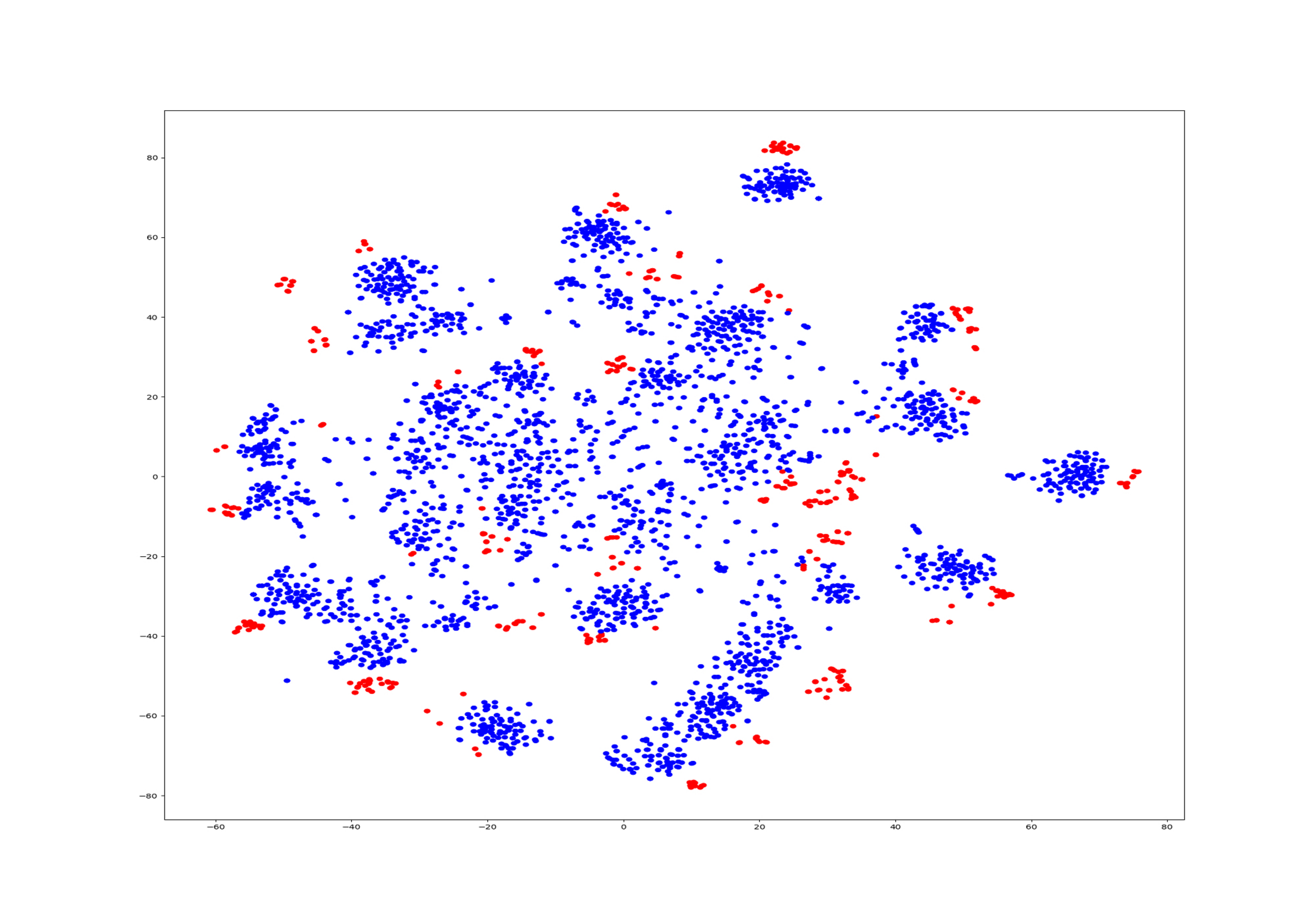}\label{fig:DANN_Src_Trg}}
    \subfigure[PCDA (D\(\rightarrow\)A)]{\includegraphics[width=0.24\textwidth]{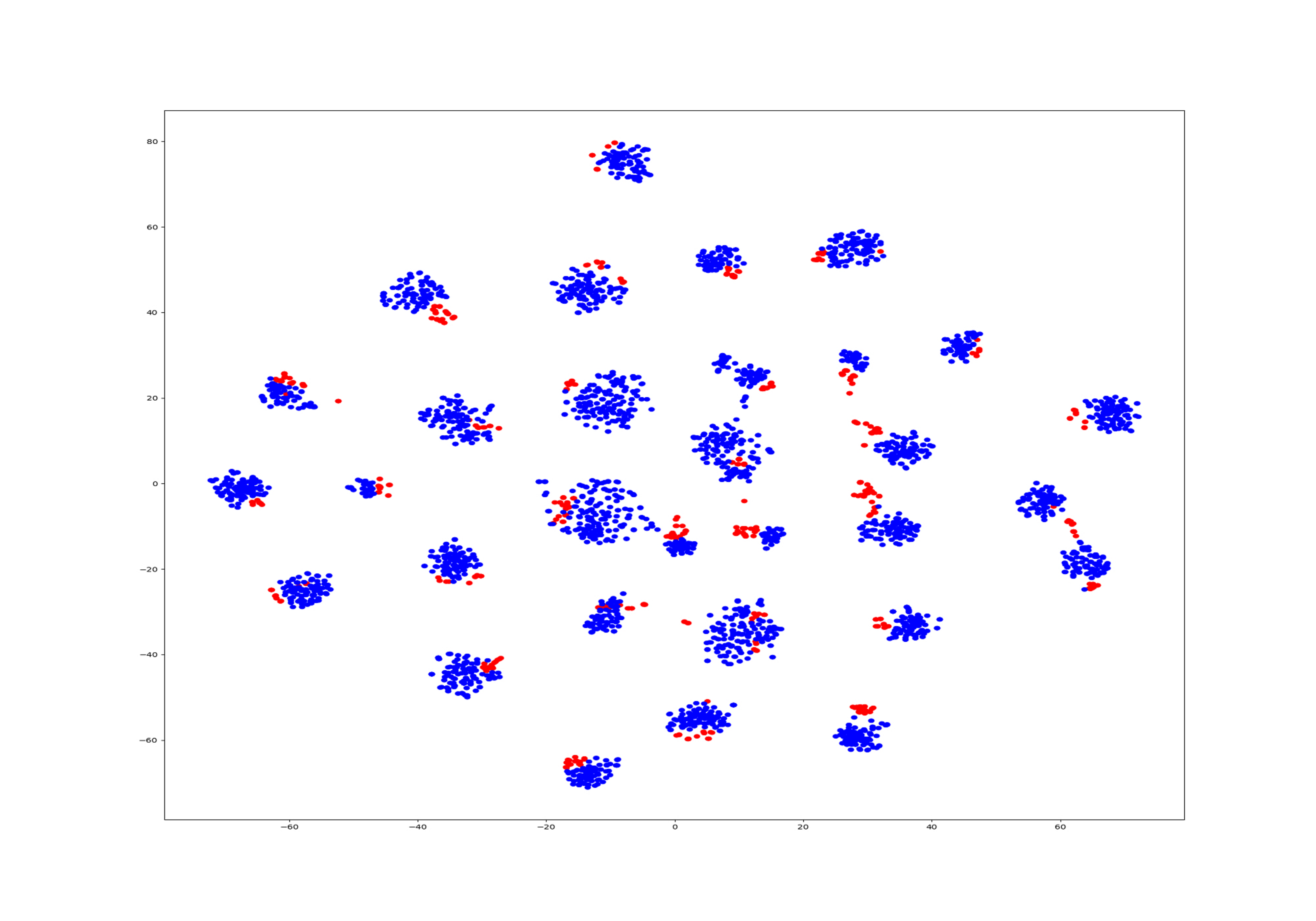}\label{fig:PCDA_Src_Trg}}
    \subfigure[Test accuracy (\(W\rightarrow A\))]{\includegraphics[width=0.24\textwidth]{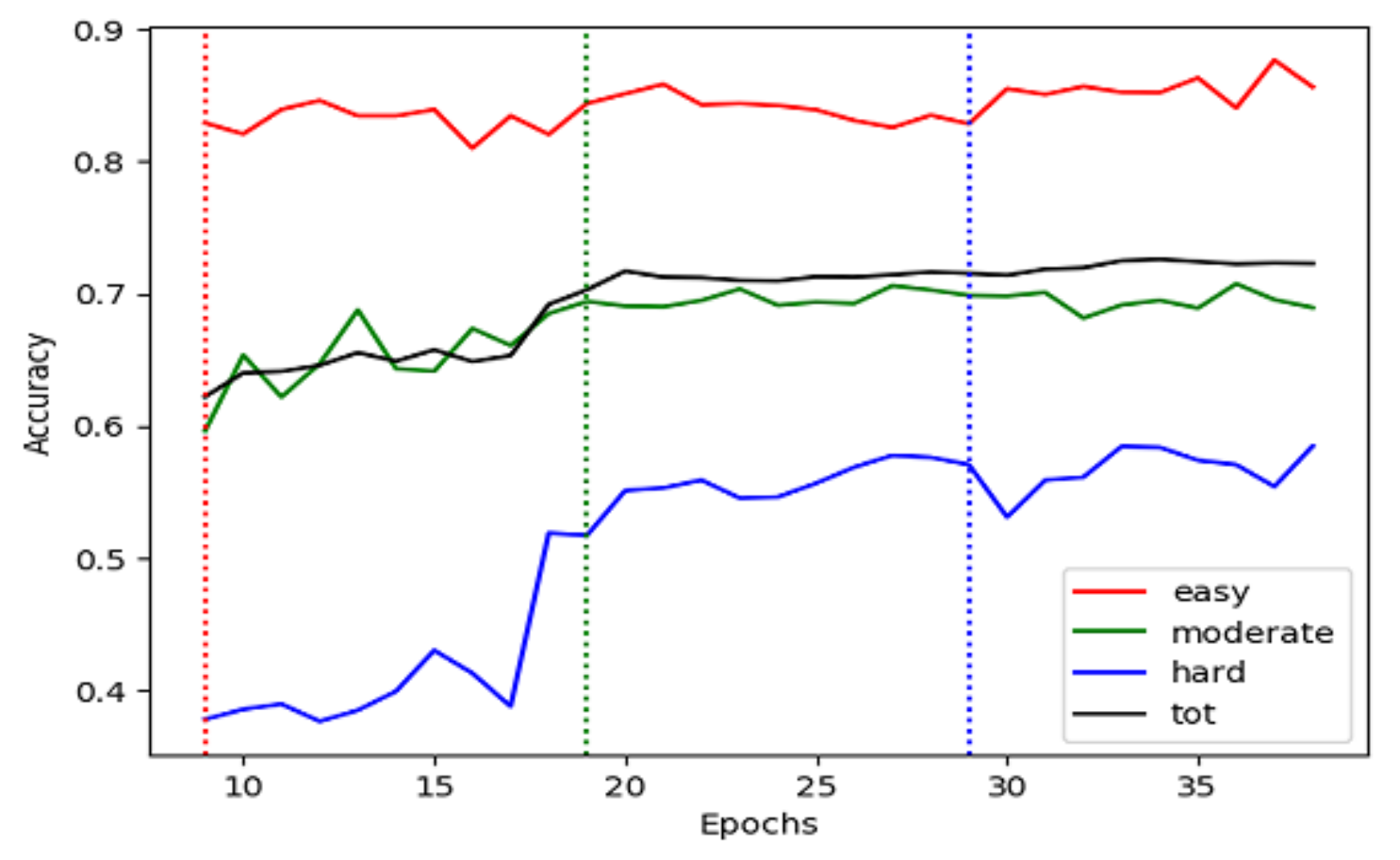}\label{fig:W_A_acc}}
    \subfigure[Test accuracy (\(D\rightarrow A\))]{\includegraphics[width=0.24\textwidth]{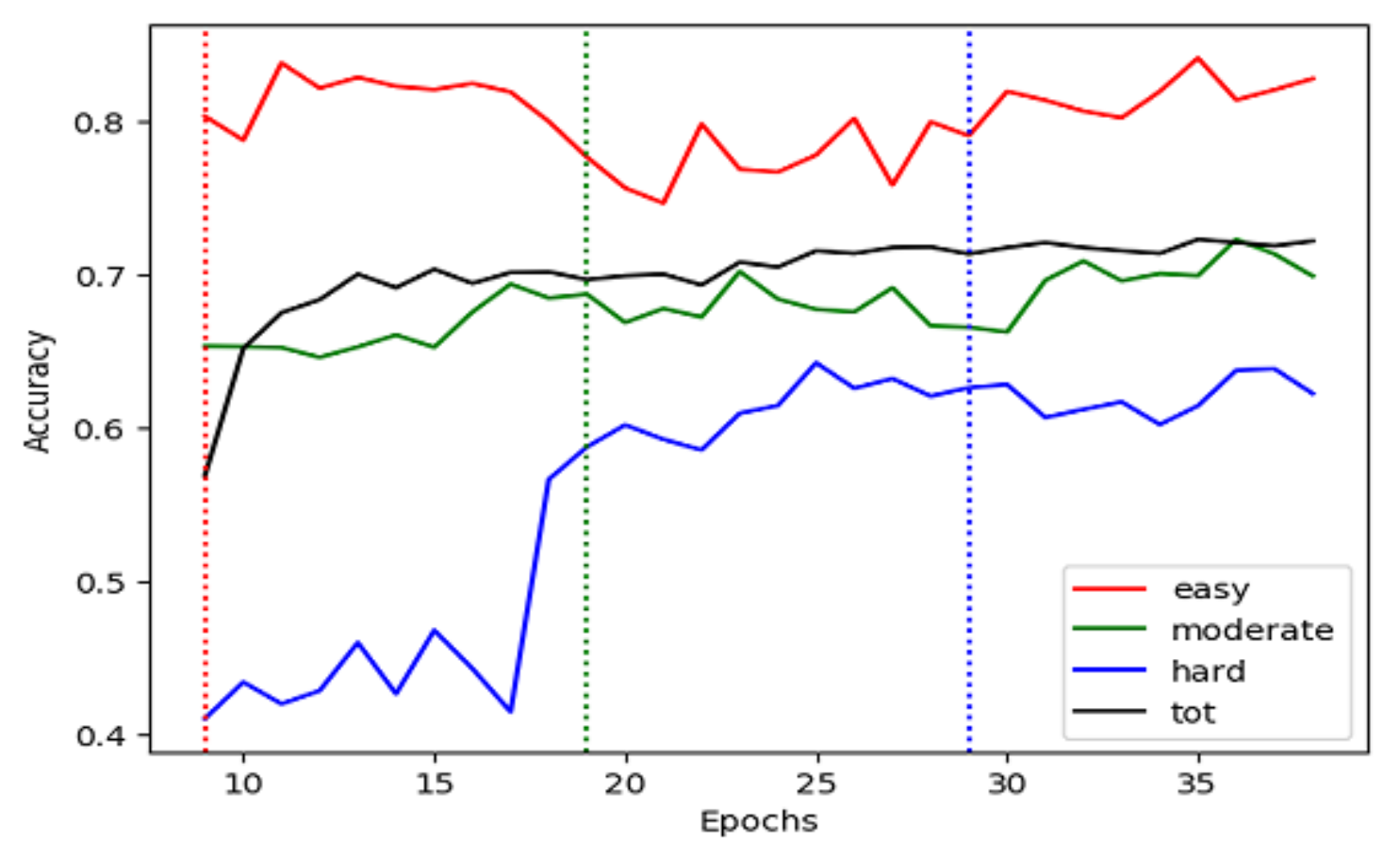}\label{fig:D_A_acc}}
    \vskip 0.1in
    \caption{On \(D\rightarrow A\) task, \textbf{(a)} and \textbf{(b)} are T-SNE visualization of features. Red points are source samples and blue are target samples. \textbf{(c)} and \textbf{(d)} shows the comparison of test accuracy and pseudo-labeling accuracy. Red line, blue and green are the pseudo-labeling accuracy of easy, moderate and hard target samples at every epoch. Black line is test accuracy of PCDA over the training epochs. The red vertical dotted line indicates the training epoch when the target data subset \(D_{e}\) is added for training our model. The green and blue vertical dotted lines are training epochs for adding \(D_{m}\) and \(D_{h}\).}
\end{figure*}
\noindent\textbf{Feature Visualization.}
To have an intuitive understanding of PCDA for domain adaptation, we visualize the target representations of the easy, moderate, and hard samples from the feature extractor \(G_{f}\) using t-SNE embedding \cite{maaten2008visualizing} on the difficult \textbf{D\(\rightarrow\)A} task. In Fig.\:\ref{fig:Easy}, we can observe that the features of the easy samples in each category are tightly clustered and distinguishable from samples in the different categories. In contrast, those of the hard samples in each category are dispersed in Fig.\:\ref{fig:Hard}. Since PCDA is developed from DANN, it is necessary to confirm whether the proposed model generates target discriminative features compared to DANN. We visualize the both source and target features of DANN and PCDA in Fig.\:\ref{fig:DANN_Src_Trg} and Fig.\:\ref{fig:PCDA_Src_Trg}. Compared with the features obtained by DANN (Fig.\:\ref{fig:DANN_Src_Trg}), Fig.\:\ref{fig:PCDA_Src_Trg} indicates that PCDA learns more category-level discriminative features. The result corroborates the efficacy of our approach.


\noindent\textbf{Effectiveness of Curriculum.}
To verify the effectiveness of pseudo-labeling curriculum, we evaluate three models for the comparison. We further investigated \textbf{Model-1} trained by only using the easy data subset \(D_{e}\), \textbf{Model-2} trained by using \(D_{e}\) and the moderate data subset \(D_{m}\) with a 2-subset curriculum, and \textbf{Model-3} trained by using \(D_{e}\), \(D_{m}\), and \(D_{h}\) with a 3-subset curriculum. \textbf{Model-3} is equivalent to our proposed PCDA. In Table \ref{table:CurriculumLearningStrategy}, \textbf{Model-3} outperforms other methods through most tasks. Although \(D_{h}\) is more likely to have false pseudo-labels, utilizing \(D_{h}\) is helpful for improving the test accuracies in most cases. It implies that the pseudo-labeling curriculum suppresses the adverse effects of the target samples with false pseudo-labels in most tasks. This table proves that our proposed method benefits the performance. However, regarding some challenging tasks in the \textit{Office-Home} dataset such as \(Rw \rightarrow Cl\) or \(Pr \rightarrow Ar\), the usage of \(D_{h}\) slightly degrades the performance due to the excessive number of false-labeled target samples. Thus, if the target specific network \(C_{t}\) still provides too many false pseudo-labels for \(D_{h}\), the effect of the pseudo-labeling curriculum degrades.    

We report the test accuracy and pseudo-labeling accuracy in Fig.\:\ref{fig:W_A_acc} and Fig.\:\ref{fig:D_A_acc}. The pseudo-labeling accuracy indicates whether pseudo-labels are assigned correctly. The test accuracy of our method is increasing when we give new additional target samples with pseudo-labels such as \(D_{m}\) or \(D_{h}\). These accord with our previous results in Table \ref{table:CurriculumLearningStrategy} in that supplementing target samples with generated pseudo-labels is effectual to boost the performance. We observe that the reliability of the generated pseudo-labels is improved iteratively as training proceeds. This proves that our training scheme with pseudo-labeling curriculum progressively improves the capability of our network to generate pseudo-labels and leads to better predictions. Moreover, the pseudo-labeling accuracies of easy samples are higher than those of both the moderate and hard samples in Fig.\:\ref{fig:W_A_acc} and Fig.\:\ref{fig:D_A_acc}. Also, the pseudo-labeling accuracies of the moderate samples are higher than those of hard samples. These experimental results provide support for our main assumption that target samples with high density values are highly likely to have correct pseudo-labels. We confirm that this assumption is valid for all our experiments including Fig.\:\ref{fig:W_A_acc} and Fig.\:\ref{fig:D_A_acc}.


\section{Conclusion}
We have proposed a novel learning scheme for unsupervised domain adaptation with the pseudo-labeling curriculum. Different from the existing works that exploit pseudo-labeled target samples, our method is robust to the adverse effect of false pseudo-labeled target samples due to the pseudo-labeling curriculum. The proposed method enables our model to generate reliable pseudo-labels progressively, and then the target specific network is able to learn target discriminative features. Furthermore, we introduce the clustering constraint on creating target features to improve the performance.  Our approach surpasses the state-of-the-art classification results on three benchmarks. We also provide extensive analysis to validate the effectiveness of our method by utilizing visualization results and various ablation studies.


{
\bibliography{main}

\begin{thebibliography}{39}
\providecommand{\natexlab}[1]{#1}
\providecommand{\url}[1]{\texttt{#1}}
\expandafter\ifx\csname urlstyle\endcsname\relax
  \providecommand{\doi}[1]{doi: #1}\else
  \providecommand{\doi}{doi: \begingroup \urlstyle{rm}\Url}\fi

\bibitem[Ben-David et~al.(2010)Ben-David, Blitzer, Crammer, Kulesza, Pereira,
  and Vaughan]{ben2010theory}
Shai Ben-David, John Blitzer, Koby Crammer, Alex Kulesza, Fernando Pereira, and
  Jennifer~Wortman Vaughan.
\newblock A theory of learning from different domains.
\newblock \emph{Machine learning}, 79\penalty0 (1-2):\penalty0 151--175, 2010.

\bibitem[Bengio et~al.(2009)Bengio, Louradour, Collobert, and
  Weston]{bengio2009curriculum}
Yoshua Bengio, J{\'e}r{\^o}me Louradour, Ronan Collobert, and Jason Weston.
\newblock Curriculum learning.
\newblock In \emph{Proceedings of the 26th annual international conference on
  machine learning}, pages 41--48. ACM, 2009.

\bibitem[Bousmalis et~al.(2016)Bousmalis, Trigeorgis, Silberman, Krishnan, and
  Erhan]{bousmalis2016domain}
Konstantinos Bousmalis, George Trigeorgis, Nathan Silberman, Dilip Krishnan,
  and Dumitru Erhan.
\newblock Domain separation networks.
\newblock In \emph{Advances in Neural Information Processing Systems}, pages
  343--351, 2016.

\bibitem[Chen et~al.(2018)Chen, Li, Sakaridis, Dai, and
  Van~Gool]{chen2018domain}
Yuhua Chen, Wen Li, Christos Sakaridis, Dengxin Dai, and Luc Van~Gool.
\newblock Domain adaptive faster r-cnn for object detection in the wild.
\newblock In \emph{Proceedings of the IEEE conference on computer vision and
  pattern recognition}, pages 3339--3348, 2018.

\bibitem[Chopra et~al.(2005)Chopra, Hadsell, and LeCun]{chopra2005learning}
Sumit Chopra, Raia Hadsell, and Yann LeCun.
\newblock Learning a similarity metric discriminatively, with application to
  face verification.
\newblock In \emph{Computer Vision and Pattern Recognition, 2005. CVPR 2005.
  IEEE Computer Society Conference on}, volume~1, pages 539--546. IEEE, 2005.

\bibitem[French et~al.(2018)French, Mackiewicz, and
  Fisher]{french2018selfensembling}
Geoff French, Michal Mackiewicz, and Mark Fisher.
\newblock Self-ensembling for visual domain adaptation.
\newblock In \emph{International Conference on Learning Representations}, 2018.
\newblock URL \url{https://openreview.net/forum?id=rkpoTaxA-}.

\bibitem[Ganin and Lempitsky(2015)]{Ganin2015}
Yaroslav Ganin and Victor Lempitsky.
\newblock Unsupervised domain adaptation by backpropagation.
\newblock In \emph{International Conference on Machine Learning}, pages
  1180--1189, 2015.

\bibitem[Goodfellow et~al.(2014)Goodfellow, Pouget-Abadie, Mirza, Xu,
  Warde-Farley, Ozair, Courville, and Bengio]{goodfellow2014generative}
Ian Goodfellow, Jean Pouget-Abadie, Mehdi Mirza, Bing Xu, David Warde-Farley,
  Sherjil Ozair, Aaron Courville, and Yoshua Bengio.
\newblock Generative adversarial nets.
\newblock In \emph{Advances in neural information processing systems}, pages
  2672--2680, 2014.

\bibitem[Guo et~al.(2018)Guo, Huang, Zhang, Zhuang, Dong, Scott, and
  Huang]{guo2018curriculumnet}
Sheng Guo, Weilin Huang, Haozhi Zhang, Chenfan Zhuang, Dengke Dong, Matthew~R.
  Scott, and Dinglong Huang.
\newblock Curriculumnet: Weakly supervised learning from large-scale web
  images.
\newblock In \emph{The European Conference on Computer Vision (ECCV)},
  September 2018.

\bibitem[He et~al.(2016)He, Zhang, Ren, and Sun]{he2016deep}
Kaiming He, Xiangyu Zhang, Shaoqing Ren, and Jian Sun.
\newblock Deep residual learning for image recognition.
\newblock In \emph{Proceedings of the IEEE conference on computer vision and
  pattern recognition}, pages 770--778, 2016.

\bibitem[Hoffman et~al.(2016)Hoffman, Wang, Yu, and Darrell]{hoffman2016fcns}
Judy Hoffman, Dequan Wang, Fisher Yu, and Trevor Darrell.
\newblock Fcns in the wild: Pixel-level adversarial and constraint-based
  adaptation.
\newblock \emph{arXiv preprint arXiv:1612.02649}, 2016.

\bibitem[Jiang et~al.(2018)Jiang, Zhou, Leung, Li, and
  Fei-Fei]{jiang2018mentornet}
Lu~Jiang, Zhengyuan Zhou, Thomas Leung, Li-Jia Li, and Li~Fei-Fei.
\newblock Mentornet: Learning data-driven curriculum for very deep neural
  networks on corrupted labels.
\newblock In \emph{International Conference on Machine Learning}, pages
  2309--2318, 2018.

\bibitem[Kim et~al.(2019)Kim, Jeong, Kim, Choi, and Kim]{Kim_2019_CVPR}
Taekyung Kim, Minki Jeong, Seunghyeon Kim, Seokeon Choi, and Changick Kim.
\newblock Diversify and match: A domain adaptive representation learning
  paradigm for object detection.
\newblock In \emph{The IEEE Conference on Computer Vision and Pattern
  Recognition (CVPR)}, June 2019.

\bibitem[Kumar et~al.(2010)Kumar, Packer, and Koller]{kumar2010self}
M~Pawan Kumar, Benjamin Packer, and Daphne Koller.
\newblock Self-paced learning for latent variable models.
\newblock In \emph{Advances in Neural Information Processing Systems}, pages
  1189--1197, 2010.

\bibitem[Lee(2013)]{lee2013pseudo}
Dong-Hyun Lee.
\newblock Pseudo-label: The simple and efficient semi-supervised learning
  method for deep neural networks.
\newblock In \emph{Workshop on Challenges in Representation Learning, ICML},
  volume~3, page~2, 2013.

\bibitem[Lin et~al.(2018)Lin, Wang, Meng, Zuo, and Zhang]{lin2018active}
Liang Lin, Keze Wang, Deyu Meng, Wangmeng Zuo, and Lei Zhang.
\newblock Active self-paced learning for cost-effective and progressive face
  identification.
\newblock \emph{IEEE transactions on pattern analysis and machine
  intelligence}, 40\penalty0 (1):\penalty0 7--19, 2018.

\bibitem[Long et~al.(2015)Long, Cao, Wang, and Jordan]{pmlr-v37-long15}
Mingsheng Long, Yue Cao, Jianmin Wang, and Michael Jordan.
\newblock Learning transferable features with deep adaptation networks.
\newblock In \emph{Proceedings of the 32nd International Conference on Machine
  Learning}, pages 97--105, 2015.

\bibitem[Long et~al.(2016)Long, Zhu, Wang, and Jordan]{long2016unsupervised}
Mingsheng Long, Han Zhu, Jianmin Wang, and Michael~I Jordan.
\newblock Unsupervised domain adaptation with residual transfer networks.
\newblock In \emph{Advances in Neural Information Processing Systems}, pages
  136--144, 2016.

\bibitem[Long et~al.(2017)Long, Zhu, Wang, and Jordan]{pmlr-v70-long17a}
Mingsheng Long, Han Zhu, Jianmin Wang, and Michael~I. Jordan.
\newblock Deep transfer learning with joint adaptation networks.
\newblock In \emph{Proceedings of the 34th International Conference on Machine
  Learning}, pages 2208--2217, 2017.

\bibitem[Long et~al.(2018)Long, CAO, Wang, and Jordan]{NIPS2018_7436}
Mingsheng Long, ZHANGJIE CAO, Jianmin Wang, and Michael~I Jordan.
\newblock Conditional adversarial domain adaptation.
\newblock In S.~Bengio, H.~Wallach, H.~Larochelle, K.~Grauman, N.~Cesa-Bianchi,
  and R.~Garnett, editors, \emph{Advances in Neural Information Processing
  Systems 31}, pages 1647--1657. Curran Associates, Inc., 2018.

\bibitem[Maaten and Hinton(2008)]{maaten2008visualizing}
Laurens van~der Maaten and Geoffrey Hinton.
\newblock Visualizing data using t-sne.
\newblock \emph{Journal of machine learning research}, 9\penalty0
  (Nov):\penalty0 2579--2605, 2008.

\bibitem[Nath~Kundu et~al.(2018)Nath~Kundu, Krishna~Uppala, Pahuja, and
  Venkatesh~Babu]{nath2018adadepth}
Jogendra Nath~Kundu, Phani Krishna~Uppala, Anuj Pahuja, and R~Venkatesh~Babu.
\newblock Adadepth: Unsupervised content congruent adaptation for depth
  estimation.
\newblock In \emph{Proceedings of the IEEE Conference on Computer Vision and
  Pattern Recognition}, pages 2656--2665, 2018.

\bibitem[Pei et~al.(2018)Pei, Cao, Long, and Wang]{pei2018multi}
Zhongyi Pei, Zhangjie Cao, Mingsheng Long, and Jianmin Wang.
\newblock Multi-adversarial domain adaptation.
\newblock In \emph{AAAI Conference on Artificial Intelligence}, 2018.

\bibitem[Russakovsky et~al.(2015)Russakovsky, Deng, Su, Krause, Satheesh, Ma,
  Huang, Karpathy, Khosla, Bernstein, et~al.]{russakovsky2015imagenet}
Olga Russakovsky, Jia Deng, Hao Su, Jonathan Krause, Sanjeev Satheesh, Sean Ma,
  Zhiheng Huang, Andrej Karpathy, Aditya Khosla, Michael Bernstein, et~al.
\newblock Imagenet large scale visual recognition challenge.
\newblock \emph{International journal of computer vision}, 115\penalty0
  (3):\penalty0 211--252, 2015.

\bibitem[Saenko et~al.(2010)Saenko, Kulis, Fritz, and
  Darrell]{saenko2010adapting}
Kate Saenko, Brian Kulis, Mario Fritz, and Trevor Darrell.
\newblock Adapting visual category models to new domains.
\newblock In \emph{European conference on computer vision}, pages 213--226.
  Springer, 2010.

\bibitem[Saito et~al.(2017)Saito, Ushiku, and Harada]{saito2017asymmetric}
Kuniaki Saito, Yoshitaka Ushiku, and Tatsuya Harada.
\newblock Asymmetric tri-training for unsupervised domain adaptation.
\newblock In \emph{International Conference on Machine Learning}, pages
  2988--2997, 2017.

\bibitem[Saito et~al.(2018)Saito, Watanabe, Ushiku, and
  Harada]{Saito_2018_CVPR}
Kuniaki Saito, Kohei Watanabe, Yoshitaka Ushiku, and Tatsuya Harada.
\newblock Maximum classifier discrepancy for unsupervised domain adaptation.
\newblock In \emph{The IEEE Conference on Computer Vision and Pattern
  Recognition (CVPR)}, June 2018.

\bibitem[Sankaranarayanan et~al.(2018)Sankaranarayanan, Balaji, Castillo, and
  Chellappa]{Sankaranarayanan_2018_CVPR}
Swami Sankaranarayanan, Yogesh Balaji, Carlos~D. Castillo, and Rama Chellappa.
\newblock Generate to adapt: Aligning domains using generative adversarial
  networks.
\newblock In \emph{The IEEE Conference on Computer Vision and Pattern
  Recognition (CVPR)}, June 2018.

\bibitem[Schroff et~al.(2015)Schroff, Kalenichenko, and
  Philbin]{schroff2015facenet}
Florian Schroff, Dmitry Kalenichenko, and James Philbin.
\newblock Facenet: A unified embedding for face recognition and clustering.
\newblock In \emph{Proceedings of the IEEE conference on computer vision and
  pattern recognition}, pages 815--823, 2015.

\bibitem[Sener et~al.(2016)Sener, Song, Saxena, and
  Savarese]{sener2016learning}
Ozan Sener, Hyun~Oh Song, Ashutosh Saxena, and Silvio Savarese.
\newblock Learning transferrable representations for unsupervised domain
  adaptation.
\newblock In \emph{Advances in Neural Information Processing Systems}, pages
  2110--2118, 2016.

\bibitem[Shimodaira(2000)]{shimodaira2000}
Hidetoshi Shimodaira.
\newblock Improving predictive inference under covariate shift by weighting the
  log-likelihood function.
\newblock \emph{Journal of statistical planning and inference}, 90\penalty0
  (2):\penalty0 227--244, 2000.

\bibitem[Shu et~al.(2018)Shu, Bui, Narui, and Ermon]{shu2018a}
Rui Shu, Hung Bui, Hirokazu Narui, and Stefano Ermon.
\newblock A {DIRT}-t approach to unsupervised domain adaptation.
\newblock In \emph{International Conference on Learning Representations}, 2018.

\bibitem[Srivastava et~al.(2014)Srivastava, Hinton, Krizhevsky, Sutskever, and
  Salakhutdinov]{srivastava2014dropout}
Nitish Srivastava, Geoffrey Hinton, Alex Krizhevsky, Ilya Sutskever, and Ruslan
  Salakhutdinov.
\newblock Dropout: a simple way to prevent neural networks from overfitting.
\newblock \emph{The Journal of Machine Learning Research}, 15\penalty0
  (1):\penalty0 1929--1958, 2014.

\bibitem[Tarvainen and Valpola(2017)]{tarvainen2017mean}
Antti Tarvainen and Harri Valpola.
\newblock Mean teachers are better role models: Weight-averaged consistency
  targets improve semi-supervised deep learning results.
\newblock In \emph{Advances in neural information processing systems}, pages
  1195--1204, 2017.

\bibitem[Tsai et~al.(2018)Tsai, Hung, Schulter, Sohn, Yang, and
  Chandraker]{tsai2018learning}
Yi-Hsuan Tsai, Wei-Chih Hung, Samuel Schulter, Kihyuk Sohn, Ming-Hsuan Yang,
  and Manmohan Chandraker.
\newblock Learning to adapt structured output space for semantic segmentation.
\newblock In \emph{Proceedings of the IEEE Conference on Computer Vision and
  Pattern Recognition}, pages 7472--7481, 2018.

\bibitem[Tzeng et~al.(2017)Tzeng, Hoffman, Saenko, and
  Darrell]{tzeng2017adversarial}
Eric Tzeng, Judy Hoffman, Kate Saenko, and Trevor Darrell.
\newblock Adversarial discriminative domain adaptation.
\newblock In \emph{Computer Vision and Pattern Recognition (CVPR)}, volume~1,
  page~4, 2017.

\bibitem[Venkateswara et~al.(2017)Venkateswara, Eusebio, Chakraborty, and
  Panchanathan]{venkateswara2017deep}
Hemanth Venkateswara, Jose Eusebio, Shayok Chakraborty, and Sethuraman
  Panchanathan.
\newblock Deep hashing network for unsupervised domain adaptation.
\newblock In \emph{Computer Vision and Pattern Recognition (CVPR), 2017 IEEE
  Conference on}, pages 5385--5394. IEEE, 2017.

\bibitem[Xie et~al.(2018)Xie, Zheng, Chen, and Chen]{xie2018learning}
Shaoan Xie, Zibin Zheng, Liang Chen, and Chuan Chen.
\newblock Learning semantic representations for unsupervised domain adaptation.
\newblock In \emph{International Conference on Machine Learning}, pages
  5419--5428, 2018.

\bibitem[Zhang et~al.(2018)Zhang, Ouyang, Li, and Xu]{zhang2018collaborative}
Weichen Zhang, Wanli Ouyang, Wen Li, and Dong Xu.
\newblock Collaborative and adversarial network for unsupervised domain
  adaptation.
\newblock In \emph{Proceedings of the IEEE Conference on Computer Vision and
  Pattern Recognition}, pages 3801--3809, 2018.

\end{thebibliography}
}
\end{document}